%% file: main.tex
\documentclass[11pt]{article}
\usepackage[numbers,sort&compress]{natbib}
\input{preamble}

\graphicspath{{png/}}

\title{
    Self-Attention at Constant Cost per Token \\
    via Symmetry-Aware Taylor Approximation
}
\author[1]{Franz A. Heinsen}
\author[2]{Leo Kozachkov}
\affil[1]{GlassRoom Software LLC}
\affil[2]{Brown University}
\date{\today}

\begin{document}

\maketitle

\begin{abstract}
The most widely used artificial intelligence (AI) models today are Transformers employing self-attention. In its standard form, self-attention incurs costs that increase with context length, driving demand for storage, compute, and energy that is now outstripping society’s ability to provide them. To help address this issue, we show that self-attention is efficiently computable to arbitrary precision with constant cost per token, achieving orders-of-magnitude reductions in memory use and computation. We derive our formulation by decomposing the conventional formulation's Taylor expansion into expressions over symmetric chains of tensor products. We exploit their symmetry to obtain feed-forward transformations that efficiently map queries and keys to coordinates in a minimal polynomial-kernel feature basis. Notably, cost is fixed inversely in proportion to head size, enabling application over a greater number of heads per token than otherwise feasible. We implement our formulation and empirically validate its correctness.\footnote{
    Source code and replication instructions are at \url{https://github.com/glassroom/sata_attention}.
} Our work enables unbounded token generation at modest fixed cost, substantially reducing the infrastructure and energy demands of large-scale Transformer models. The mathematical techniques we introduce are of independent interest.
\end{abstract}

\section{Introduction}

Most artificial intelligence (AI) services today are applications of Transformer models employing self-attention, a mathematical mechanism for capturing sequential dependencies over tokens \cite{turner2024introductiontransformers, vaswani2023attentionneed}. In its conventional formulation, self-attention has $\bigO(n)$ space and time complexity per token, where $n$ is the number of tokens in the sequence, or context. The memory and compute required to process each additional token increase in proportion to the number of tokens in the sequence. The time complexity for the sequence as a whole is quadratic, $\bigO(n^2)$. All else remaining the same, processing each new token in a context requires more computing infrastructure. End-users, who benefit from longer personalized context histories and step-by-step reasoning chains, seek out AI services that can successfully handle an ever greater number of tokens. With global adoption of AI services increasing rapidly, demand for new data centers and energy sources is outstripping society's current ability to provide them \cite{stanford2025aiindexreport}.

Numerous modifications and replacements for conventional self-attention have been proposed to address its ever-rising memory and compute requirements. Proposed modifications include data and parameter reuse schemes, local and structured context windows, and low-rank and sparse approximations \cite{tay2022efficient, weng2020transformerfamily, weng2023transformerfamily2}. The proposed replacements are mainly recurrent neural networks (RNNs) with linear recurrences computable via parallel scan \cite{blelloch1990a, blelloch1990b}, including state space models (SSMs) \cite{arora2024simple, feng2024rnnsneeded, gu2024mamba, gu2022efficientlymodelinglongsequences, katharopoulos2020transformers}.

\begin{figure}[t]
	\begin{center}
		\includegraphics[width=\textwidth]{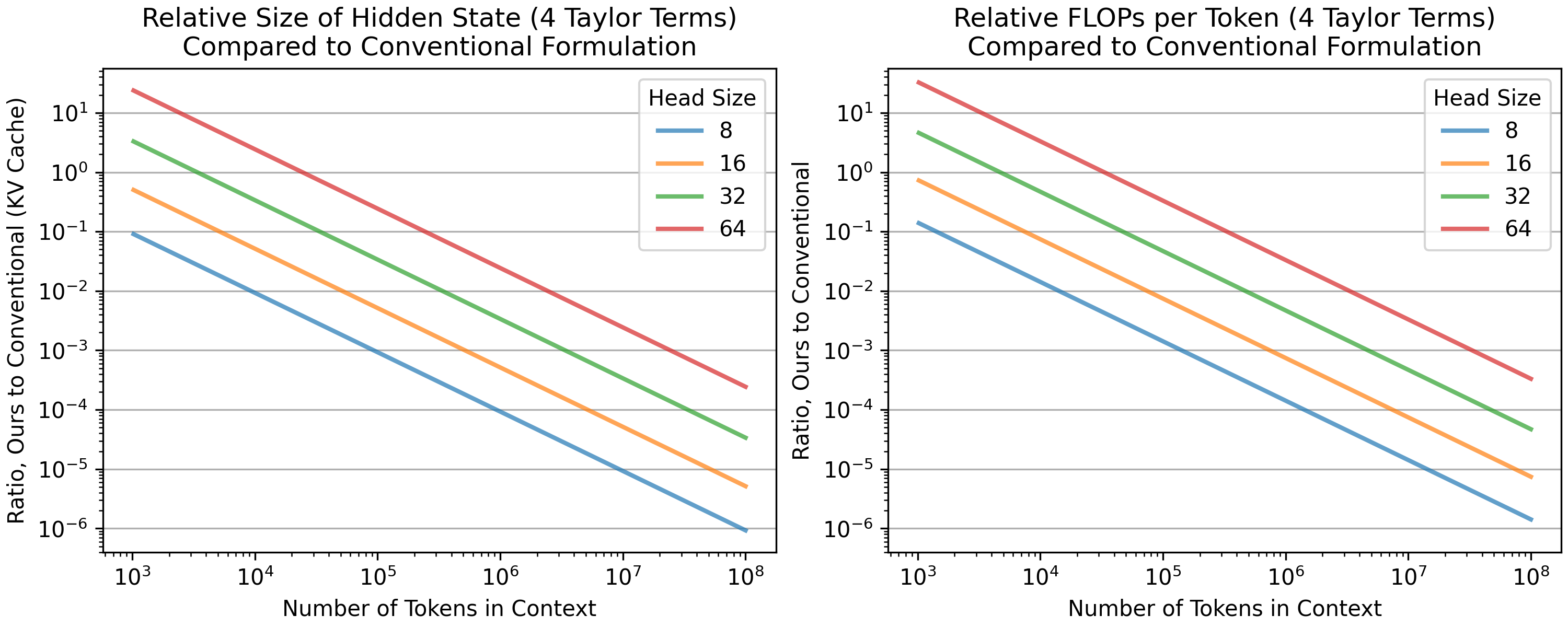}
		\caption{Hidden state size and FLOPs per token (4 Taylor terms), compared to conventional formulation.}
		\label{fig:hidden_state_size_and_flops_vs_conventional}
	\end{center}
\end{figure}

Rather than propose another modification or replacement, we show that self-attention can be efficiently computed at any desired precision at constant cost per token, $\bigO(1)$, with a hidden state of size
\begin{equation}
\left( \valSize + 1 \right)
\binom{\keySize + P - 1}{P - 1},
\quad \textcomment{// fixed number of elements in hidden state}
\end{equation}

and a constant number of floating-point operations (FLOPs) per token in the forward pass,
\begin{equation}
\left(
4 \valSize
+
\frac{2\left(P \, \keySize + 1\right)}{\keySize + 1}
+
2
\right)
\binom{\keySize + P - 1}{P - 1},
\quad \textcomment{// fixed FLOPs per token in forward pass}
\end{equation}

where $\keySize$ and $\valSize$ are key and value sizes, respectively, and $P$ is the number of terms in a Taylor series expansion, controlling precision. In practice, we find that four Taylor terms ($P = 4$) suffice for recovering conventional attention with elementwise errors of approximately the same magnitude as Float16 resolution, acceptable for many AI applications. As we increase the number of tokens in context, our formulation's costs per token become orders of magnitude more efficient than previously possible (Figure \ref{fig:hidden_state_size_and_flops_vs_conventional}). Our formulation is a form of linear attention \cite{katharopoulos2020transformers}, computable via parallel scan. 

In comparison, the conventional formulation has a hidden state of variable size, the key--value cache (``KV cache,'' for short), with $n (\keySize + \valSize)$ elements \cite{pope2022efficientlyscalingtransformerinference}, where $n$ denotes the number of tokens in context, executes a variable number of FLOPs per token in the forward pass, $n (2 \keySize + 2 \valSize + 3)$  \cite{hoffmann2022trainingcomputeoptimallargelanguage}, and cannot be computed via parallel scan, because its stepwise transformation is not associative over tokens.

Our formulation enables Transformer models to generate tokens in perpetuity with modest fixed costs per token, at any desired precision, so long as the Transformer's positional encoding scheme can accommodate sequences of indefinite length. New levers for trading off memory use, compute cost, and numerical precision become available. The number of Taylor terms, $P$, controls precision. Space and time complexity are inversely proportional to $\keySize$ and $\valSize$, such that we can reduce the fixed per-token costs by making heads smaller and applying self-attention over a greater number of heads. Alternatively, if given a fixed budget, we can apply self-attention over a greater number of heads per token than otherwise feasible. In contrast, with the conventional formulation, space and time complexity are proportional to the number of heads, because each head's per-token costs are proportional to context length.

Previous efforts to approximate self-attention via Taylor expansion have stopped at the quadratic term ({\em i.e.}, second order) due to the perceived complexity of evaluating all necessary polynomial interactions for higher-degree terms \cite{arora2024simple,qiu2023mbtaylorformer, dass2023vitality, xu2024qt, nauen2025taylorshift}. We show that the Taylor expansion decomposes into an expression over symmetric chains of tensor products, and their symmetric structure naturally reveals the minimal basis for all polynomial interactions. Our key contribution is a maximally succinct, computationally efficient, and embarrassingly parallel feed-forward transformation that evaluates the associated kernel function to arbitrary Taylor truncation order at constant cost per token.

We implement our formulation, verify that it recovers attention with increasing accuracy as we expand the Taylor series, and confirm that it reduces memory use and run time by orders of magnitude as we increase context length, compared to the conventional formulation. Our results likely understate the achievable savings in optimized implementations. All evidence indicates that our work here can substantially reduce the infrastructure and energy demands of large-scale Transformer models.

\section{Deriving Our Formulation}

\subsection{Taylor Expansion of Core Operation}

The core operation in self-attention consists of the dot-product of a query and a key vector, scaling the result, and applying the exponential function to it. Given query and key vectors $q, k \in \mathbb{R}^{\keySize}$ and a conventional constant $c = \sqrt{\keySize}$, the core operation's Taylor expansion is:
\begin{equation}\label{eq:infinite_taylor_expansion}
\exp \left( \frac{q^\top k}{c} \right)
=
\sum_{p=0}^{\infty} \, \alpha_p \, (q^\top k)^p,
\qquad
\alpha_p \coloneq \frac{1}{p!\, c^p}.
\\
\end{equation}

\subsection{Decomposing into Expressions over Symmetric Tensors}

For each integer $p > 0$ in \eqref{eq:infinite_taylor_expansion}, we can express $\big( q^\top k \big)^p$ as:
\begin{equation}\label{eq:chains_of_tensor_products}
{\small\thinmuskip=1mu\medmuskip=2mu\thickmuskip=3mu
\begin{aligned}
\left( q^\top k \right)^1
& = q^\top k
&& = \sum_{i=1}^{\keySize} q_i k_i
&& = \sum \underbrace{
	q \odot k
}_{\textstyle \in \mathbb{R}^{\keySize}}
\\[-0.5em]
(q^\top k)^2
& = \sum_{i_1=1}^{\keySize} q_{i_1} k_{i_1} \sum_{i_2=1}^{\keySize} q_{i_2} k_{i_2}
&& = \sum_{i_1=1}^{\keySize} \sum_{i_2=1}^{\keySize} (q_{i_1} q_{i_2}) (k_{i_1} k_{i_2})
&& = \sum \underbrace{
	\left(q \otimes q\right) \odot (k \otimes k)
}_{\textstyle  \in \mathbb{R}^{\keySize \times \keySize} }
\\[-0.5em]
(q^\top k)^3
& = \sum_{i_1=1}^{\keySize} q_{i_1} k_{i_1} \sum_{i_2=1}^{\keySize} q_{i_2} k_{i_2} \sum_{i_3=1}^{\keySize} q_{i_3} k_{i_3}
&& = \sum_{i_1=1}^{\keySize} \sum_{i_2=1}^{\keySize} \sum_{i_3=1}^{\keySize} (q_{i_1} q_{i_2} q_{i_3}) (k_{i_1} k_{i_2} k_{i_3})
&& = \sum \underbrace{
	(q \otimes q \otimes q) \odot (k \otimes k \otimes k)
}_{\textstyle  \in \mathbb{R}^{\keySize \times \keySize \times \keySize} }, \\
\vdots \quad
& \quad\quad \vdots
&& \quad\quad \vdots
&& \quad\quad \vdots
\end{aligned}
}
\end{equation}

\begin{figure}[t]
	\begin{center}
		\includegraphics[width=\textwidth]{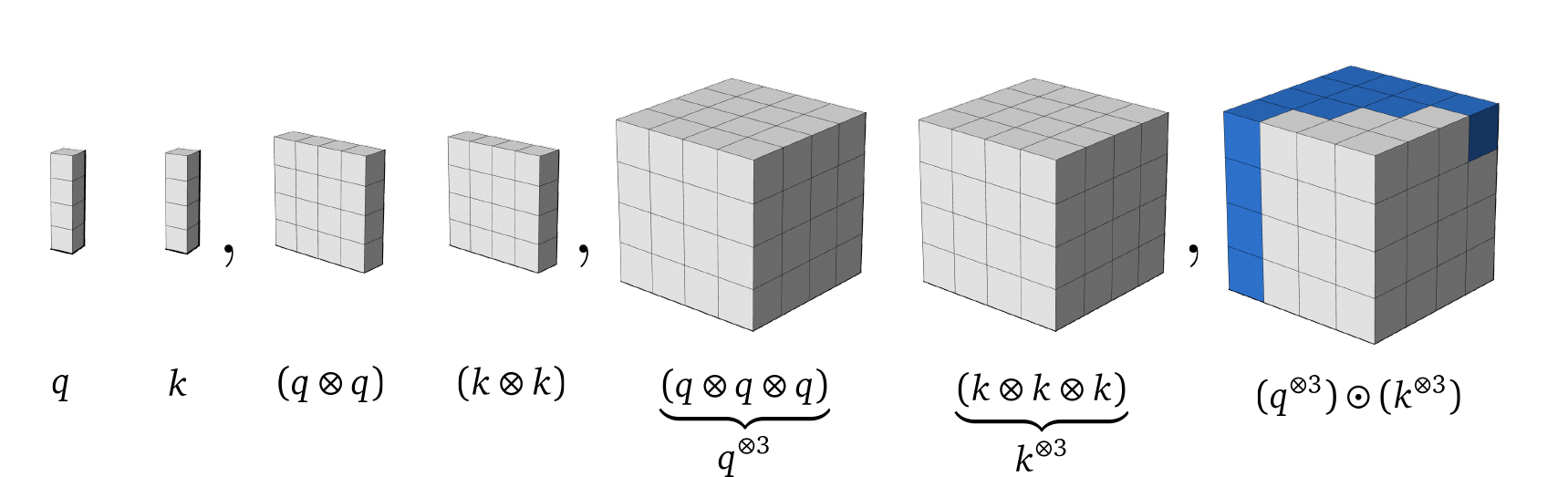}
		\caption{Illustration of the symmetric tensors of increasing order in \eqref{eq:chains_of_tensor_products}. The blue region in the rightmost plot contains the minimal monomial basis for representing $\big( q^\top k \big)^{p}$, as discussed in Section \ref{sec:minimal_basis}.}
		\label{fig:chains_of_tensor_products}
	\end{center}
\end{figure}

where $\odot$ denotes elementwise (Hadamard) product, and $\otimes$ denotes tensor (outer) product. See Figure \ref{fig:chains_of_tensor_products} for a helpful illustration. Generalizing, we obtain the following expression for $p > 0$:
\begin{equation}\label{eq:dot_product_power_is_sum_of_symmetric_tensor}
\left( q^\top k \right)^p
\quad = \sum_{i_1=1}^{\keySize} \sum_{i_2=1}^{\keySize} \dots \sum_{i_p=1}^{\keySize} (q_{i_1} q_{i_2} \dots q_{i_p}) (k_{i_1} k_{i_2} \dots k_{i_p})
\quad = \sum (q^{\otimes p}) \odot (k^{\otimes p}),
\end{equation}

where $q^{\otimes p}$ and $k^{\otimes p}$ are order-$p$ symmetric tensors obtained by chains of tensor products:
\begin{equation}
	\begin{aligned}
		q^{\otimes p}
		& = \underbrace{
			q \otimes q \otimes \dots \otimes q
		}_{p \textcomment{ identical factors}}
		\\
		k^{\otimes p}
		& = \underbrace{
			k \otimes k \otimes \dots \otimes k
		}_{p \textcomment{ identical factors}}.
	\end{aligned}
\end{equation}

By construction, the elements of $q^{\otimes p}$ and $k^{\otimes p}$ are monomials multiplying all possible combinations of $p$ scalar elements of $k$ and $q$, respectively. The number of all such possible combinations is $\keySize^p$.

\subsection{Identifying the Minimal Basis in Each Symmetric Tensor}\label{sec:minimal_basis}

The elementwise multiplication of the two symmetric tensors, $(q^{\otimes p}) \odot (k^{\otimes p})$, in \eqref{eq:dot_product_power_is_sum_of_symmetric_tensor}, evaluates to a third symmetric tensor, $(q \odot k)^{\otimes p}$, whose $\keySize^p$ elements are the monomials that sum to $\big( q^\top k \big)^p$ (Figure \ref{fig:chains_of_tensor_products}). Each of these three order-$p$ tensors being symmetric, its upper hyper-triangular region, consisting of
\begin{equation}
m_p = \binom{\keySize + p - 1}{p}
\end{equation}

elements indexed by $i_1 \le i_2 \le \dots \le i_p$, contains the tensor's unique elements \cite{Schatz2014tensors, kostrikin1989linear}. All monomials outside that region are permutations of a monomial in the region. Therefore, the upper hyper-triangular regions of $q^{\otimes p}$ and $k^{\otimes p}$ contain the $m_p$ unique monomials combining elements of $q$ and $k$, respectively, that make up the {\em minimal basis} in the space of such monomials, for obtaining $\big( q^\top k \big)^p$. 

To obtain the final correct expression for $\big( q^\top k \big)^p$, the elements of this minimal basis must be weighted appropriately before being summed. A simple example with $\qrySize = p =  2$ helps to illustrate this point. Letting $q = (q_1,q_2)$ and $k = (k_1,k_2)$:
\[
(q^\top k)^2
=
\left(q_1 k_1 + q_2 k_2\right)^2
=
q_1^2 k_1^2
+ q_1 q_2 k_1 k_2
+ q_2 q_1 k_2 k_1
+ q_2^2 k_2^2.
\]
The middle two terms are identical, and can therefore be combined into a single term $2 q_1 q_2 k_1 k_2$. In general, we can distinguish the monomial \textit{basis components} from the monomial \textit{coefficients}. In the example above, $q_1 q_2 k_1 k_2$ is a basis component and $2$ is its coefficient. 

In summary, the expansion of $(q^\top k)^p$ can be written as a sum over the $m_p$ distinct monomials corresponding to the upper hyper-triangular region of $(q \odot k)^{\otimes p}$, where each monomial appears with a multiplicity determined by the number of index permutations that produce it. This separation between a minimal set of basis monomials and their associated combinatorial coefficients will allow us to reformulate $(q^\top k)^p$ as a structured inner product.

\subsection{From Symmetric Tensors to a Weighted Feature Inner Product}
It follows from the previous subsection that the expansion of $(q^\top k)^p$ can be organized by grouping identical monomials and weighting each by its multiplicity. This naturally suggests defining a feature map $\Phi_p(\cdot)$ whose components correspond to the $m_p$ basis monomials, together with a diagonal weighting that accounts for their combinatorial coefficients. With these definitions, $(q^\top k)^p$ can be written as a weighted inner product,
\begin{equation}\label{eq:weighted_inner_product}
\left( q^\top k \right)^p
=
\left< \Phi_p(q), \Phi_p(k) \right>_{C_p},
\end{equation}

where $\Phi_p(q)$ and $\Phi_p(k)$ denote application of a feature map $\Phi_p : \mathbb{R}^{\keySize} \to \mathbb{R}^{m_p}$ that obtains the $m_p$ monomials in the upper hyper-triangular region of $q^{\otimes p}$ and $k^{\otimes p}$, respectively, {\em tightly packed in a vector}; and $C_p \in \mathbb{R}^{m_p \times m_p}$ is a constant diagonal weight matrix, each diagonal element of which is the number of permutations in the full symmetric tensor corresponding to each unique monomial in the tightly packed vector, scaling that unique monomial by its frequency in the full symmetric tensor.

A well-known result from kernel theory \citep{shawe2004kernel} is that for each integer $p \ge 0$, there exists a reproducing kernel Hilbert space (RKHS) $\mathcal{H}_p$ and an associated feature map, such that degree-$p$ monomials can be expressed as an inner product in $\mathcal{H}_p$. So far, we have ignored $p = 0$. We incorporate it by treating $q^{\otimes 0}$ and $k^{\otimes 0}$ as empty chains of tensor products that evaluate to scalar value $1$, as is conventional, and obtain a conventional RKHS feature map, with $\Phi_0(q) = \Phi_0(k)= 1$. See Appendix~\ref{appsec:incorporating_degree_zero} for the details.

\paragraph{Truncating for Approximation }
With this feature map perspective, the Taylor series \eqref{eq:infinite_taylor_expansion} becomes an infinite sum of scaled weighted inner products. In practice, we truncate the series to a finite number of terms $P$, obtaining an approximation:
\begin{equation}\label{eq:truncated_taylor_expansion_with_inner_products}
	\exp \left( \frac{q^\top k}{c} \right)
	\approx
	\sum_{p=0}^{P-1} \, \alpha_p \, \left< \Phi_p(q), \Phi_p(k) \right>_{C_p}.
\end{equation}

\paragraph{Efficient Computation}
For each power $p$, the weight matrix, $C_p$, is {\em constant}, so the elements in its diagonal can be precomputed in advance, only once, for subsequent efficient weighting of the inner product via $m_p$ elementwise multiplications. The indices that select an order-$p$ upper hyper-triangular region are also {\em constant}, and therefore can be similarly precomputed in advance, only once, for subsequent efficient computation of $\Phi_p(\cdot)$ via parallel multiplication of elements selected from its input vector:
\begin{equation}
\Phi_p(x)
\coloneqq
\setlength{\arraycolsep}{0em}
\left[
	\;
	\displaystyle
	\prod_{j=1}^{p} x_{\scriptscriptstyle M_{p_{\,ij}}}
	\quad \scriptstyle i = 1, 2, \dots, m_p
	\; 
\right],
\end{equation}

where $M_p \in \{1, 2, \dots, \keySize \}^{m_p \times p}$ is the constant matrix with precomputed indices. In pseudo-code:
\begin{equation}\label{eq:phi_x_pseudo_code}
\boxed{
    \begin{array}{l}
    \\[-0.5em]
    \; \text{\tt \small def Phi(x): return x[..., M].prod(dim=-1)} \; \\
    \\[-0.5em]
    \end{array}
} \; ,
\end{equation}

a maximally succinct, computationally efficient, embarrassingly parallel transformation. The matrix $M_p$, each row containing indices $i_1, i_2, \dots, i_p$, has a hierarchical structure, given by $i_1 \le i_2 \le \dots \le i_p$, opening additional opportunities for improving computational efficiency, which we leave for future work.\footnote{It's not too hard to show that for $p > 0$, every monomial indexed by $M_{p-1}$ is a factor of a monomial indexed by $M_p$.}

\begin{figure}[t]
	\begin{center}
		\includegraphics[width=\textwidth]{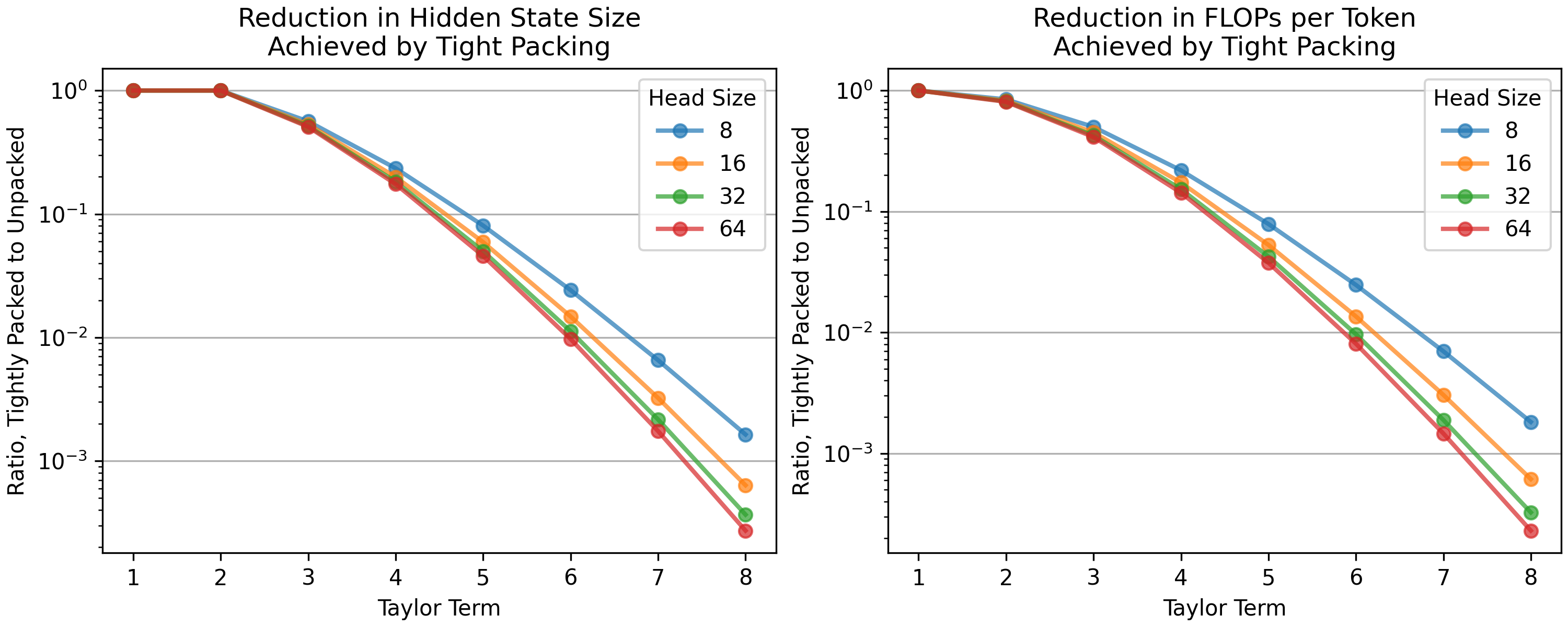}
		\caption{Reduction in hidden state size and FLOPs per token achieved by our tight packing method.}
		\label{fig:reduction_achieved_by_tight_packing}
	\end{center}
\end{figure}

By grouping identical monomials, tightly packing them into vectors, and weighting each unique monomial by its multiplicity, our method reduces the number of features we must store and the number of floating-point operations (FLOPs) we must execute, per token, {\em by orders of magnitude}, compared to a feature map that naively evaluates all possible permutations of the unique monomials (Figure \ref{fig:reduction_achieved_by_tight_packing}).

\subsection{Linear Attention with Our Formulation}\label{ssec:applying_as_linear_attention}
So far, we have described how the exponentiated dot product can be approximated as a weighted sum of inner products. Compared to conventional Linear Transformer formulations (e.g., \citep{katharopoulos2020transformers,choromanski2020rethinking}), the key distinction is that our approximation consists of multiple inner products (one per term in the truncated Taylor expansion) rather than a single inner product. As a result, whereas standard Linear Transformers can be evaluated using a single parallel prefix scan, our approach requires $P$ such scans, corresponding to the $P$ Taylor terms. Importantly however, these scans are independent of each other and can be executed in parallel, after which their outputs are aggregated to form the final approximation.

Please see Appendix~\ref{appsec:linear_attention} for the details.

\subsection{Hidden State Size}

We define ``hidden state size'' as the number of elements in accumulated state, per token, for linear attention (Appendix~\ref{appsec:linear_attention}). For one Taylor term with degree $p$, hidden state size is:
\begin{equation}
    ( \valSize + 1 ) \binom{\keySize + p - 1}{p}.
    \quad \textcomment{// hidden state size, one Taylor term}
\end{equation}

For all Taylor terms in a model, hidden state size is $\displaystyle \sum_{p=0}^{P-1} \left( ( \valSize + 1 ) \binom{\keySize + p - 1}{p} \right)$, equal to:
\begin{equation}
    ( \valSize + 1 )  \binom{\keySize + P - 1}{P - 1}.
    \quad \textcomment{// hidden state size, all Taylor terms}
\end{equation}

Compared to a naive feature map that computes $\keySize^p$ monomials for a Taylor term of degree $p$, our formulation reduces hidden state size by orders of magnitude (Figure \ref{fig:reduction_achieved_by_tight_packing}, left panel). The conventional formulation of attention's equivalent to a hidden state is the KV cache, consisting of $n (\keySize + \valSize)$ elements \cite{pope2022efficientlyscalingtransformerinference}, where $n$ denotes the number of tokens in context. The left panel of Figure \ref{fig:hidden_state_size_and_flops_vs_conventional} compares our formulation's hidden state size (for 4 Taylor terms) to the size of the conventional formulation's KV cache, for different head sizes and context lengths.

\subsection{FLOPs per Token}

We define floating-point operations (FLOPS) per token as the number of multiply and add operations that must be executed in the forward pass to compute attention for the token. Adding up the number of those operations, FLOPs per token for one Taylor term with degree $p$ is:
\begin{equation}
    (4 \valSize + 2p + 4) \binom{\keySize + p - 1}{p}.
    \quad \textcomment{// FLOPs per token, one Taylor term}
\end{equation}

For all Taylor terms in a model, FLOPs per token is $\displaystyle \sum_{p=0}^{P-1} \left( (4 \valSize + 2p + 4) \binom{\keySize + p - 1}{p} \right)$, equal to:
\begin{equation}
\left(
4 \valSize
+
\frac{2\left(P \, \keySize + 1\right)}{\keySize + 1}
+
2
\right)
\binom{\keySize + P - 1}{P - 1}.
\quad \textcomment{// fixed FLOPs per token, all Taylor terms}
\end{equation}

Compared to a naive feature map that computes $\keySize^p$ monomials for a Taylor term of degree $p$, our formulation reduces FLOPs per token by orders of magnitude (Figure \ref{fig:reduction_achieved_by_tight_packing}, right panel). The conventional formulation of attention executes $n (2 \keySize + 2 \valSize + 3)$ FLOPS per token in the forward pass, where $n$ denotes the number of tokens in context.\footnote{
	We measure forward-pass FLOPs per token for the conventional formulation with the same method as \cite{hoffmann2022trainingcomputeoptimallargelanguage}: (a) FLOPs for computing the $QK^T$ logits, equal to $2 \times n_\mathrm{queries} \times n_\mathrm{tokens} \times \keySize \times n_\mathrm{heads}$; plus (b) FLOPs for applying a Softmax function over each row of scores, equal to $3 \times n_\mathrm{queries} \times n_\mathrm{tokens}$; plus (c) FLOPs for contracting the attention matrix with values, equal to $2 \times n_\mathrm{queries} \times n_\mathrm{tokens} \times \valSize \times n_\mathrm{heads}$. We specify $n_\mathrm{queries} = 1$ and $n_\mathrm{heads} = 1$, and simplify the sum of (a), (b), and (c).
}. The right panel of Figure \ref{fig:hidden_state_size_and_flops_vs_conventional} compares our formulation's FLOPs per token in a forward pass (for 4 Taylor terms) to that of the conventional formulation, for different head sizes and context lengths.

\subsection{Costs Fixed Inversely in Proportion to Head Size}

\begin{figure}[t]
	\begin{center}
		\includegraphics[width=\textwidth]{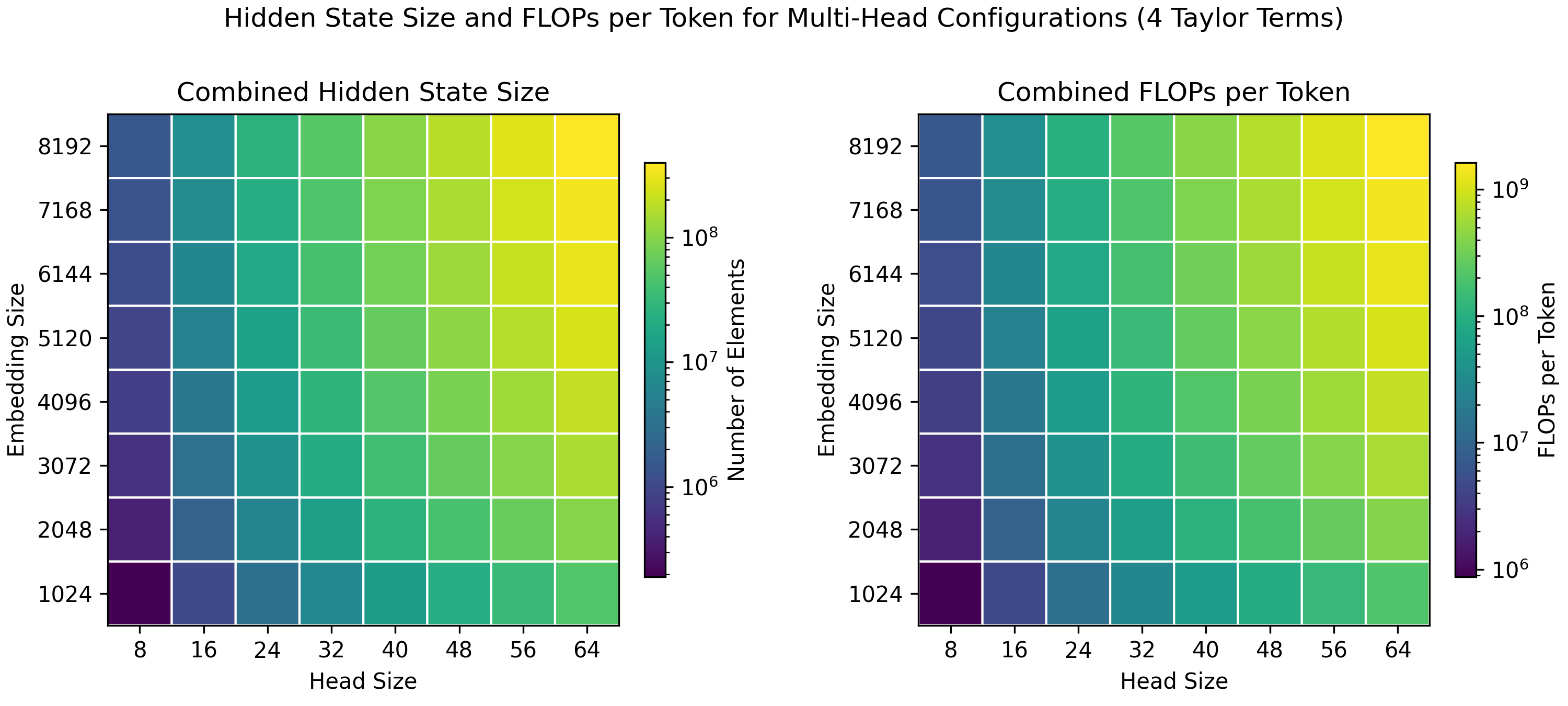}
		\caption{Size of hidden state and FLOPs per token in multi-head configurations.}
		\label{fig:hidden_sz_and_flops_for_multiple_heads}
	\end{center}
\end{figure}

Our formulation fixes costs per token inversely in proportion to $\keySize$ and $\valSize$, enabling us to reduce costs by making heads smaller and applying attention over a greater number of heads than otherwise feasible, given an embedding size (Figure \ref{fig:hidden_sz_and_flops_for_multiple_heads}). In contrast, with the conventional formulation, memory use and run time per token increase in proportion to the number of heads, because each head's per-token costs are proportional to context length, limiting the number of heads that are feasible in practice.

\section{Implementation}\label{sec:implementation}

We implement our formulation, and verify that it successfully recovers self-attention computed conventionally. We also verify that our implementation successfully reduces memory use and run time by orders of magnitude, compared to those of the conventional formulation. Our implementation relies on PyTorch, a commonly used software framework for parallel computation \cite{paszke2019pytorchimperativestylehighperformance}.

\subsection{Successful Recovery of Conventional Self-Attention}
We measure our implementation's ability to recover conventional self-attention over sequences with up to 100K tokens, and successfully verify that each additional Taylor term we add to the series decreases reconstruction error. See Appendices \ref{appsec:recovering_conventional} and \ref{appsec:recovering_by_token_position} for detailed reconstruction results.

\begin{figure}[t]
	\begin{center}
		\includegraphics[width=\textwidth]{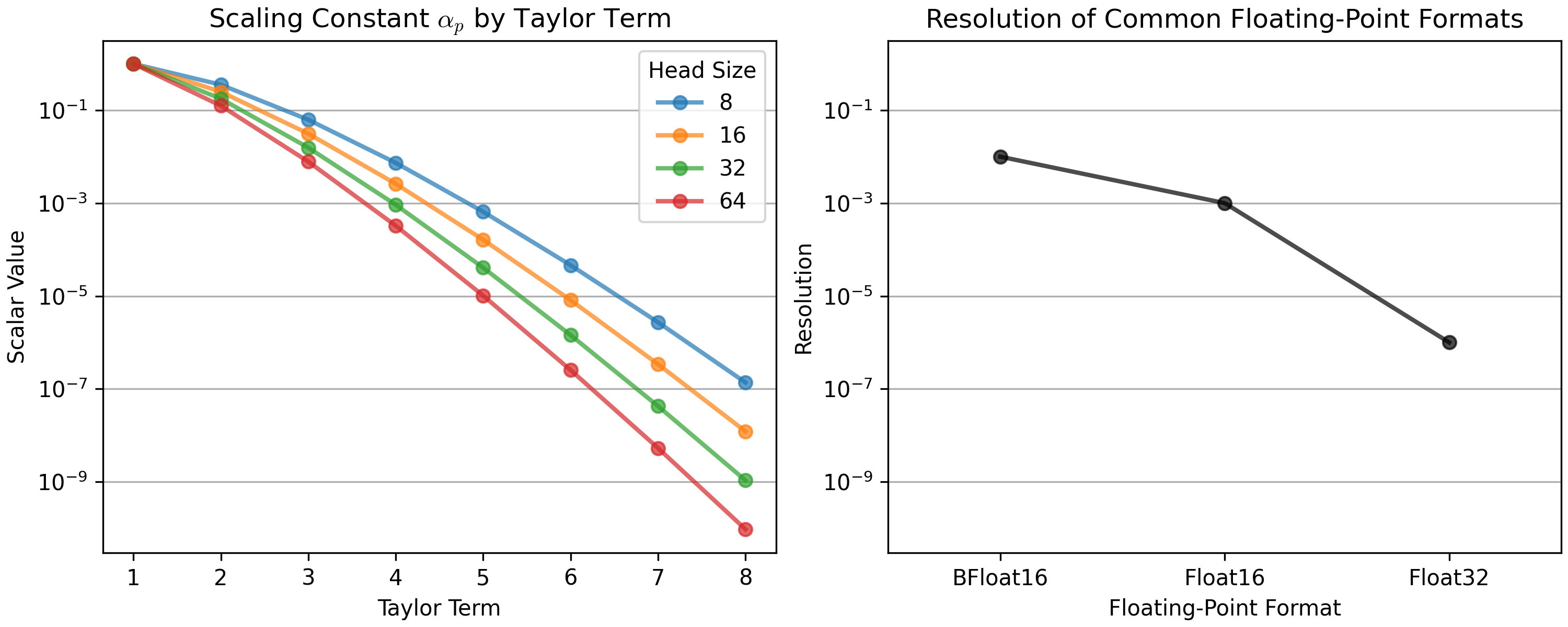}
		\caption{Scaling constant of each Taylor term, $\alpha_p$, compared to resolution of common floating-point formats.}
		\label{fig:scaling_constant_versus_float_resolution}
	\end{center}
\end{figure}

We find that four Taylor terms, $P = 4$, suffice for recovering conventional attention with elementwise errors of roughly the same magnitude as Float16 resolution, which is not too surprising, because the magnitude of scaling constant $\alpha_p$ declines to that resolution by the fourth Taylor Term (Figure \ref{fig:scaling_constant_versus_float_resolution}).

\subsection{Successful Reduction of Memory Use and Run Time by Orders of Magnitude}
We measure peak memory allocated and run time, per token, of our proof-of-concept implementation, compared to a conventional implementation, given a context that we increase in length from 1K to 100M tokens, with four different head sizes, on an Nvidia GPU. All operations are executed with autocasting, {\em i.e.}, letting PyTorch decide which floating-point format to use, as necessary, for matching the precision required by each operation. The peak memory allocation figures include temporary space consumed by interim data. Run time is measured as the mean of seven executions.

\begin{figure}[t]
	\begin{center}
		\includegraphics[width=\textwidth]{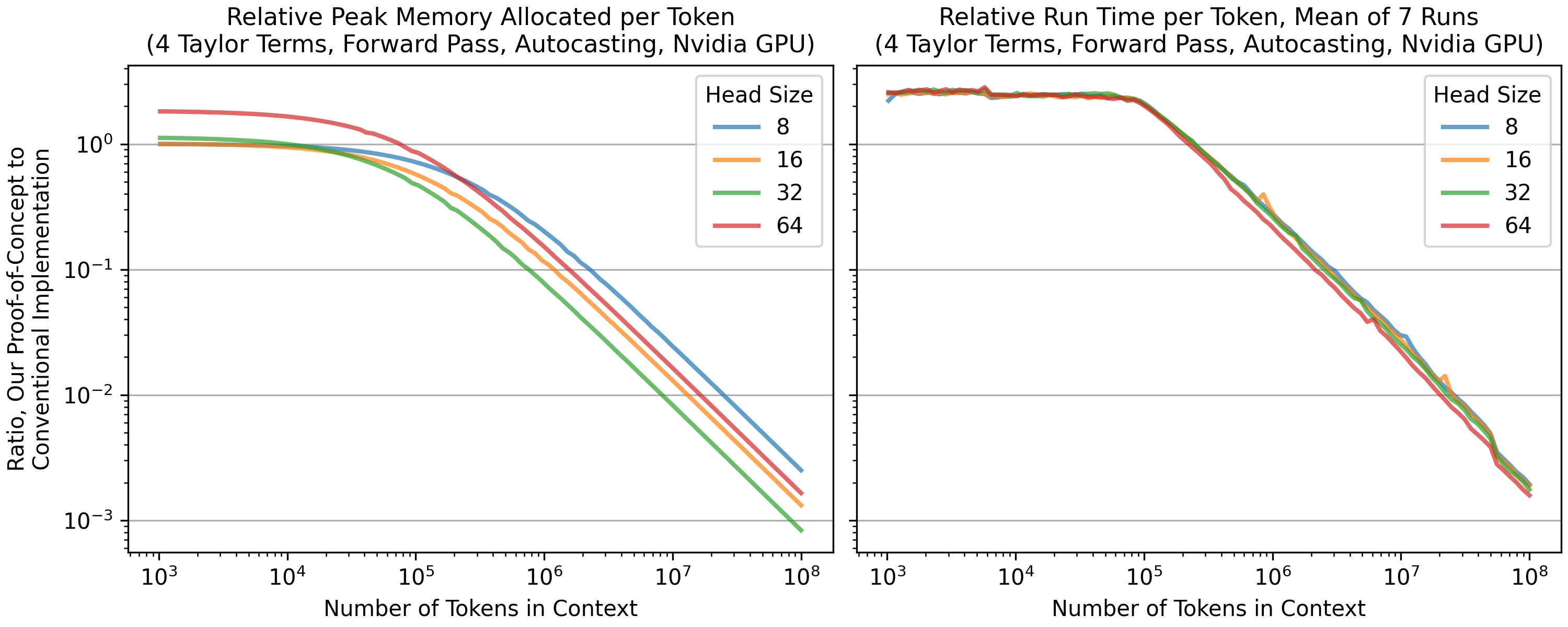}
		\caption{Our proof-of-concept implementation's memory use and run time per token, compared to those of a conventional formulation. Our implementation is unfairly penalized by the issues discussed in \ref{ssec:proof_of_concept}.}
		\label{fig:poc_benchmarks_against_conventional}
	\end{center}
\end{figure}

{\em Both memory use and run time of our implementation are unfairly penalized by the issues discussed in \ref{ssec:proof_of_concept}.} Despite these unfair disadvantages, our implementation's peak memory allocated, per token, declines three orders of magnitude below that of conventional attention as we increase context length to 100M tokens (Figure \ref{fig:poc_benchmarks_against_conventional}, left panel). Run time, per token, declines almost three orders of magnitude below conventional attention as we increase context length to 100M tokens (Figure \ref{fig:poc_benchmarks_against_conventional}, right panel).

\subsection{Proof of Concept}\label{ssec:proof_of_concept}
Our implementation is an initial one, and should properly be considered a proof of concept. Its source code consists primarily of high-level calls to PyTorch's preexisting libraries, without any customization. Unlike implementations of the conventional formulation, which has now benefited from nearly a decade of performance optimization work by a large community of AI researchers and practitioners, our formulation is new and has yet to benefit from comparable long-term work.

Targets for performance optimization include:

\paragraph{Unnecessary Temporary Copying of Query and Key Elements}
{\em Currently, we make $m_p \times p$ temporary copies of elements from each query and key vector, instead of pointing to those elements}, for mapping the vector to $m_p$ monomial features. We rely on PyTorch's advanced indexing facilities to obtain views of shape $m_p \times p$ from every vector, but each ``view'' returned by PyTorch actually makes a copy of the data, which can saturate memory bandwidth. In principle, copying all that data is unnecessary.

\paragraph{Absence of Optimizations that Exploit Hierarchical Structure of Symmetric Indices}
{\em Currently, we do not exploit the structure of the indices stored in  matrix $M_p$}. Each row of $M_p$ contains indices $i_1, i_2, \dots, i_p$ organized hierarchically by $i_1 \le i_2 \le \dots \le i_p$, opening additional opportunities for improving computational efficiency. In principle, it should be possible to exploit hierarchical structure to reduce memory and compute use.

\paragraph{Sequential Instead of Parallel Evaluation of Taylor Terms}
{\em Currently, we evaluate Taylor Terms sequentially, instead of in parallel}. We queue them on a single stream on an Nvidia GPU. As we discuss in \ref{ssec:applying_as_linear_attention}, Taylor terms can be evaluated in parallel, because they are independent of each other.

\paragraph{Absence of Common On-Device Optimizations}
{\em Currently, our focus is on validating the correctness of our formulation, not on developing a high-performance implementation}. A more efficient implementation requires writing low-level on-device code ({\em e.g.}, a CUDA kernel for Nvidia GPUs) that carefully handles data, both to avoid unnecessarily making temporary copies of it, and to ensure it is more frequently on faster-access memory ({\em e.g.}, HBM instead of SRAM on Nvidia GPUs) as needed for computation.

\paragraph{Absence of Additional Performance Optimizations}
{\em Currently, we have not explored any additional performance optimizations}. They include the possibility of reducing the dimensionality of higher-order feature spaces (say, for $P \ge 4$) by applying conventional techniques, such as dropping basis components with low-magnitude coefficients in $C_p$, finding best-fit lower-rank basis approximations, and obtaining fast approximations of the basis via random sampling or random projections.\footnote{We should add that many methods for improving the conventional formulation's space and time complexity also benefit our formulation ({\em e.g.}, data and parameter reuse schemes like sharing a single key-value head with multiple query heads).}

\section{Discussion}
We have presented a formulation of self-attention that approximates the exponential kernel underlying its Softmax function while eliminating its dependence on context length. For a fixed approximation order, both memory use and computation become constant per token. Rather than modifying the definition of attention, we reorganize its computation by approximating every exponentiated dot product with a truncated Taylor expansion and expressing the resulting terms in a compact, symmetry-aware feature representation. As the approximation order increases to infinity, our construction converges to conventional attention, allowing it to be recovered to arbitrary accuracy subject to numerical precision.

\paragraph{Implications for long-context inference}

The most immediate consequence of approximating the exponential kernel with our method is that the marginal cost of extending context no longer increases with sequence length. In conventional attention, longer contexts require proportionally larger key--value caches and proportionally more computation per generated token, increasing pressure on memory capacity, processor utilization, communications bandwidth, and energy consumption. Our formulation replaces the variable-size cache with a fixed-size accumulated state whose size and cost depend only on head dimensions and the chosen approximation order.

A second implication is architectural. Because per-token state size and computation decrease in proportion to head dimension, attention can be distributed across many smaller heads without incurring the linear growth in cost that arises in conventional multi-head attention. While we do not evaluate trained models here, the formulation makes such configurations feasible in regimes where key--value caching would otherwise dominate resource usage.

\paragraph{Accuracy--efficiency trade-offs}

Our approach introduces an explicit accuracy--efficiency trade-off through the choice of approximation order. Unlike sparsity-based or low-rank methods, which alter the attention pattern itself, our method targets the same attention pattern and approximates it directly.

In practice, we find that only a small number of terms is sufficient to match conventional attention to within typical reduced-precision numerical error. This suggests that, for many inference settings, higher-order contributions to the kernel fall below the noise floor imposed by floating-point arithmetic. Nevertheless, the appropriate approximation order may depend on model architecture, head size, and representation statistics, and should be treated as a tunable hyperparameter.

\paragraph{Relation to prior work}

The proposed method fits naturally within the family of linear attention mechanisms, in which attention can be computed via streaming accumulation rather than explicit storage of past keys and values. The main distinction is that we approximate the exponential kernel as a sum of multiple, independently accumulated components, rather than attempting to represent it with a single feature map. This perspective clarifies why previous Taylor-based approaches typically stopped at very low order: without exploiting symmetry, the number of polynomial features grows combinatorially. By identifying and tightly packing the minimal set of unique monomials, we make higher-order approximations practical at modest head sizes.

\paragraph{Limitations and Future Directions}

Our implementation is intentionally a proof of concept and leaves several questions open. First, our benchmarks isolate the attention mechanism itself and do not measure end-to-end throughput in trained models, where other components may dominate runtime. Second, our recovery experiments verify correctness using synthetic inputs, but do not evaluate downstream task performance or training dynamics.

In addition, the current implementation incurs overheads that are not inherent to the method, including temporary data copies and sequential evaluation of approximation terms. These factors penalize performance relative to what a fused, hardware-optimized implementation could achieve, and the reported runtime results should therefore be interpreted conservatively.

Although the attention mechanism itself incurs constant per-token cost for a fixed approximation order, unbounded-context generation also requires positional encoding schemes that remain well-behaved at extreme sequence lengths. Such schemes are complementary to, rather than replaced by, the proposed formulation.

Finally, head dimension plays a more prominent role than in conventional attention. The formulation is most favorable for small heads and modest approximation orders, and should be understood as opening a new region of the architectural design space rather than uniformly improving all configurations.

Several directions for future work follow naturally. An important next step is to train Transformer models, end-to-end and via transfer learning, using the proposed attention mechanism, and evaluate downstream performance and convergence behavior. Doing so at large scale will require developing fused hardware kernels that target the performance optimizations we discuss in \ref{ssec:proof_of_concept}. Further opportunities include compressing higher-order feature spaces and extending the approach to other analytic kernels beyond the exponential. More generally, our work enables exploration of new approaches to address the rising demand for storage, compute, and energy, driven by growing adoption of AI services.

\bibliography{biblio}

\clearpage
\section*{Appendix}
\addcontentsline{toc}{section}{Appendix}

\setcounter{section}{0}
\renewcommand{\thesection}{A\arabic{section}}
\renewcommand{\theHsection}{A\arabic{section}}

\section{Incorporating Degree Zero in Feature Map}\label{appsec:incorporating_degree_zero}

We incorporate the first Taylor term, with $p = 0$, by treating $q^{\otimes 0}$ and $k^{\otimes 0}$ as empty chains of tensor products that evaluate to scalar value $1$, as is conventional, obtaining:
\begin{equation}
\begin{aligned}
m_0
& = \binom{\keySize + 0 - 1}{0}
&& = 1
\\
M_0
& = m_0 \times 0 \; \text{matrix}
&& = []
\\
\Phi_0(\cdot)
& = \text{(empty product)}
&& = 1.
\end{aligned}
\end{equation}

With this change, $\Phi_p$ becomes a conventional RKHS feature map from vectors in $\mathbb{R}^{\keySize}$ to $m_p$ coordinates in the minimal basis of monomials for obtaining $\left( q^\top k \right)^p$, for all $p \ge 0$.

\section{Linear Attention}\label{appsec:linear_attention}

Consider causal self-attention over a sequence of tokens
\[
\{(q_t, k_t, v_t)\}_{t=1}^T,
\qquad
q_t, k_t \in \mathbb{R}^{\keySize},
\quad
v_t \in \mathbb{R}^{\valSize}.
\]
The causal attention output at time $T$ is
\[
y_T
=
\frac{
\sum\limits_{t=1}^{T}
\exp\!\left(\frac{q_T^\top k_t}{c}\right)\, v_t
}{
\sum\limits_{t=1}^{T}
\exp\!\left(\frac{q_T^\top k_t}{c}\right)
}.
\]

We approximate the exponential kernel using the truncated expansion \eqref{eq:truncated_taylor_expansion_with_inner_products}. Substituting this approximation into the attention expression yields
\[
y_T
\approx
\frac{
\sum\limits_{t=1}^{T}
\sum\limits_{p=0}^{P-1}
\alpha_p
\left\langle
\Phi_p(q_T), \Phi_p(k_t)
\right\rangle_{C_p}
\, v_t
}{
\sum\limits_{t=1}^{T}
\sum\limits_{p=0}^{P-1}
\alpha_p
\left\langle
\Phi_p(q_T), \Phi_p(k_t)
\right\rangle_{C_p}
}.
\]

Reordering the sums gives
\begin{equation}\label{eq:approx_kernel_self_attention}
y_T
\approx
\frac{
\sum\limits_{p=0}^{P-1}
\alpha_p
\sum\limits_{t=1}^{T}
\left\langle
\Phi_p(q_T), \Phi_p(k_t)
\right\rangle_{C_p}
\, v_t
}{
\sum\limits_{p=0}^{P-1}
\alpha_p
\sum\limits_{t=1}^{T}
\left\langle
\Phi_p(q_T), \Phi_p(k_t)
\right\rangle_{C_p}
}.
\end{equation}

\paragraph{Accumulated Feature States}

The inner product is linear in its second argument, so the denominator terms factor as
\[
\sum_{t=1}^{T}
\left\langle
\Phi_p(q_T), \Phi_p(k_t)
\right\rangle_{C_p}
=
\left\langle
\Phi_p(q_T),
\sum_{t=1}^{T} \Phi_p(k_t)
\right\rangle_{C_p} .
\]
The numerator is vector-valued due to the presence of $v_t$, motivating the
introduction of degree-wise accumulated states. For each degree $p$, define
\[
Z_{p,T}
:=
\alpha_p \sum_{t=1}^T \Phi_p(k_t),
\qquad
S_{p,T}
:=
\alpha_p \sum_{t=1}^T \Phi_p(k_t)\, v_t^\top .
\]
These states satisfy the linear recurrences
\[
Z_{p,T} = Z_{p,T-1} + \alpha_p \Phi_p(k_T),
\qquad
S_{p,T} = S_{p,T-1} + \alpha_p \Phi_p(k_T) v_T^\top .
\]
Both updates are linear and can therefore be computed efficiently via either
sequential recurrence or parallel prefix-sum (scan).

\paragraph{Evaluation at the Query}

By construction,
\[
S_{p,T}^\top h
=
\alpha_p \sum_{t=1}^T
\left\langle
h, \Phi_p(k_t)
\right\rangle_{C_p} v_t
\qquad
\text{for any } h .
\]
Evaluating at $h = \Phi_p(q_T)$ gives
\[
S_{p,T}^\top \Phi_p(q_T)
=
\alpha_p \sum_{t=1}^T
\left\langle
\Phi_p(q_T), \Phi_p(k_t)
\right\rangle_{C_p} v_t ,
\]
which matches the numerator contributions in
\eqref{eq:approx_kernel_self_attention}. Similarly,
\[
\left\langle
\Phi_p(q_T), Z_{p,T}
\right\rangle_{C_p}
=
\alpha_p \sum_{t=1}^T
\left\langle
\Phi_p(q_T), \Phi_p(k_t)
\right\rangle_{C_p} .
\]

\paragraph{Final Attention Expression}

Aggregating over all degrees $p$, define
\[
Z_T
:=
\sum_{p=0}^{P-1}
\left\langle
\Phi_p(q_T), Z_{p,T}
\right\rangle_{C_p},
\qquad
S_T
:=
\sum_{p=0}^{P-1}
S_{p,T}^\top \Phi_p(q_T).
\]
The approximate causal self-attention output can then be written compactly as
\[
y_T \approx \frac{S_T}{Z_T}.
\]

\clearpage
\section{Recovering Conventional Self-Attention}\label{appsec:recovering_conventional}

We measure our proof-of-concept implementation's ability to recover conventional self-attention with four different head sizes, $\keySize = \valSize = d$, and four different Taylor truncation numbers, $P$, over autoregressive (causal) sequences of 100K tokens, sampling each query, key, and value from $\mathcal{N}(0, 1)^{d}$. We compare the outputs of our implementation, evaluated at auto-casted precision, to those of a conventional implementation, evaluated with the highest-precision floating-point format supported by Nvidia GPUs, Float64, which we treat as ``ground truth.'' We measure elementwise differences in decimal orders of magnitude ({\em e.g.}, an absolute difference of $0.01$ is measured as $-2$ decimal orders of magnitude).

\vskip 0.2in
\begin{center}
    \includegraphics[width=\textwidth]{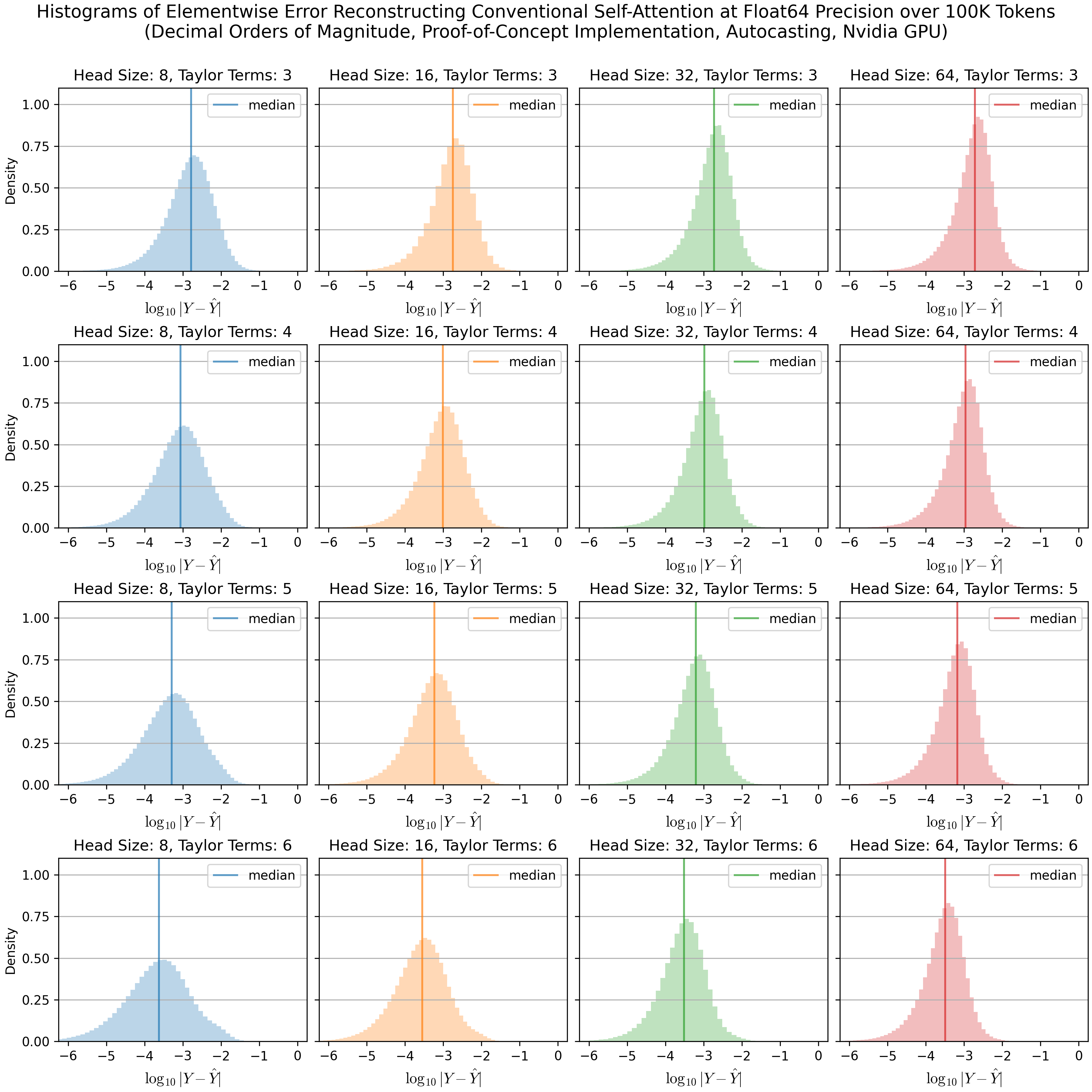}
\end{center}

\clearpage
\section{Recovering Conventional Self-Attention by Token Position}\label{appsec:recovering_by_token_position}

We measure our proof-of-concept implementation's ability to recover conventional self-attention with four different head sizes, $\keySize = \valSize = d$, and four different Taylor truncation numbers, $P$, over autoregressive (causal) sequences of 100K tokens, sampling each query, key, and value from $\mathcal{N}(0, 1)^{d}$. We compare the outputs of our implementation, evaluated at auto-casted precision, to those of a conventional implementation, evaluated with the highest-precision floating-point format supported by Nvidia GPUs, Float64, which we treat as ``ground truth.'' We measure elementwise differences in decimal orders of magnitude ({\em e.g.}, an absolute difference of $0.01$ is measured as $-2$ decimal orders of magnitude).

\vskip 0.2in
\begin{center}
    \includegraphics[width=\textwidth]{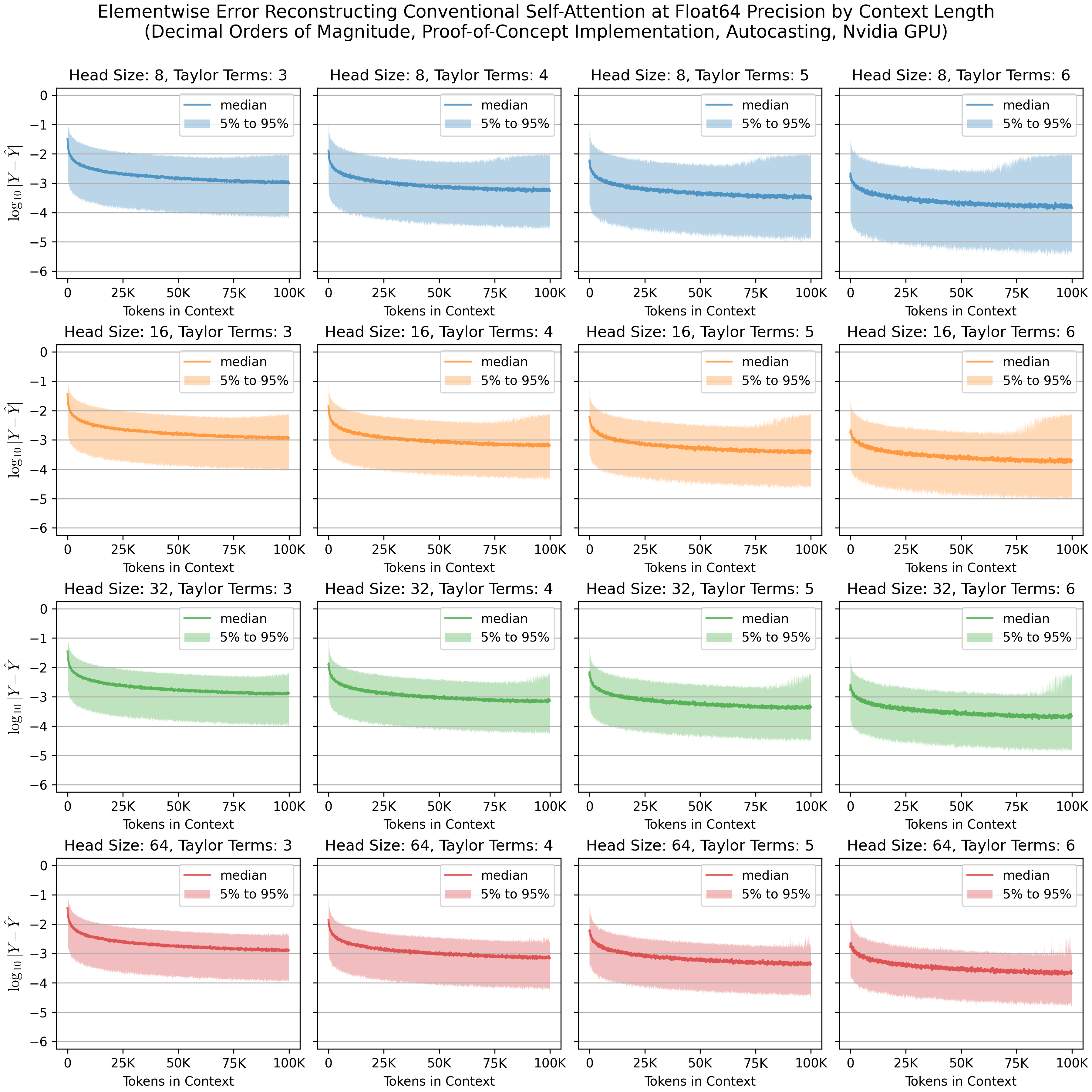}
\end{center}

\end{document}

%% file: preamble.tex
\usepackage[left=1in, right=1in, top=1.25in, bottom=1.25in]{geometry}
\usepackage[parfill]{parskip}            
\usepackage{setspace}                    
\usepackage[final, expansion=alltext]{microtype} 
\pdfoutput=1                             

\usepackage[bitstream-charter]{mathdesign}  
\usepackage[scaled=0.92]{PTSans}            
\usepackage{amsmath}                        
\usepackage{mathtools}                      
\usepackage{bbm}                            
\usepackage{accents}                        

\let\oldstar\*
\renewcommand{\*}[1]{\mathbf{#1}}

\DeclareMathOperator{\bigO}{\mathcal{O}}

\newcommand{\qrySize}{d_K}
\newcommand{\keySize}{d_K}
\newcommand{\valSize}{d_V}

\newcommand{\textcomment}[1]{\text{\scriptsize\sffamily{#1}}}

\usepackage[table, dvipsnames]{xcolor}
\definecolor{shadecolor}{gray}{0.9}
\definecolor{Gray}{gray}{0.8}
\definecolor{LightCyan}{rgb}{0.88,1,1}
\definecolor{green(html/cssgreen)}{rgb}{0.0, 0.5, 0.0}

\usepackage{graphicx}
\usepackage[labelfont={it,small}, font=small]{caption}
\usepackage[most]{tcolorbox}

\usepackage[colorlinks, linktoc=all]{hyperref}
\hypersetup{
    citecolor=MidnightBlue,
    linkcolor=black,
    urlcolor=MidnightBlue
}
\usepackage[nameinlink]{cleveref}

\usepackage{natbib}

\usepackage{titling}
\usepackage{authblk}

\usepackage{mdframed}
\newmdtheoremenv{theo}{Theorem}

\newtheorem{theorem*}{Theorem}
